\newif\ifieee
\IfFileExists{IEEEtran.cls}{\ieeetrue}{\ieeefalse}
\ifieee
  \documentclass[conference]{IEEEtran}
\else
  \documentclass[10pt,twocolumn,letterpaper]{article}
  \usepackage[margin=0.7in,columnsep=0.25in]{geometry}
  \usepackage{times}
\fi
\ifieee\else
  \newenvironment{IEEEkeywords}{\par\noindent\textit{Index Terms}---}{\par}
  \newcommand{\IEEEauthorblockN}[1]{#1}
  \newcommand{\IEEEauthorblockA}[1]{\\ #1}
\fi
\usepackage{amsmath,amssymb}
\usepackage{graphicx}
\usepackage{booktabs}
\usepackage{multirow}
\usepackage{xcolor}
\usepackage[hidelinks]{hyperref}

\begin{document}
\title{Where Mixture Recognition Fails: An Ingredient-Level Audit of Unseen Combinations in SmellNet}

\author{\IEEEauthorblockN{Zirui Liu\textsuperscript{*}}%
\IEEEauthorblockA{School of Future Science and Engineering\\Soochow University, Suzhou, China\\Email: 2662404018@stu.suda.edu.cn}%
\and
\IEEEauthorblockN{Zheming Zhu\textsuperscript{*}}%
\IEEEauthorblockA{Baibuti Innovation College\\South China University of Technology, Guangzhou, China\\\textit{*Equal contribution.}}}
\maketitle

\begin{abstract}
An electronic nose must name every ingredient of a mixture, including combinations it has never measured. SmellNet, a public benchmark for this task, scores ingredients with a hit rate that is told how many each sample contains. We audit what this protocol hides. We hold out ingredient combinations from the SmellNet training data, evaluate four architectures (MLP, CNN, LSTM and a Transformer encoder) without revealing the ingredient count, and repeat the evaluation once on the official unseen split. Under the count-informed protocol, 90--96\% of mixtures receive an ingredient that is absent, on our folds and on the official split alike. On our folds, every model identifies one ingredient reliably but the second rarely, and it completes the pair with a familiar training partner about twice as often as chance. On the official split, even the first-ranked ingredient is correct in only 29--46\% of binary mixtures, often because a frequent training ingredient that is absent from every test pair is ranked first, and on its combinations that are truly absent from training no model names both ingredients in more than 1.2\% of them. The failures fall on the same ingredients for every architecture, and a frequency correction only moves errors between ingredients. Mixture recognition is therefore limited by ingredients the models cannot separate and by a preference for frequent ingredients, not by the architecture, and it should be evaluated with set-level errors and without oracle cardinality.
\end{abstract}


\section{Introduction}

An electronic nose identifies odours from the joint response of an array of broadly selective gas sensors~\cite{persaud1982model}. With low-cost sensors, such arrays can recognise food, spices and other natural substances, but real samples are rarely pure. A practical electronic nose must therefore name every ingredient of a mixture, including mixtures whose combination of ingredients it has never measured. This is a form of compositional generalization: the ingredients are known, while their combination is new~\cite{keysers2020cfq,hupkes2020compositionality}. SmellNet~\cite{feng2026smellnet} provides a large public testbed for this setting. It records 12 food ingredients alone and in binary and ternary mixtures with low-cost gas sensors, and its unseen test split is intended to contain ingredient pairings that never occur in training.

The SmellNet authors describe the unseen split as a challenge for compositional generalization, and the models they report score far lower on it than on mixtures seen in training. Its ingredient-level metric is a dynamic Top-$K$ hit rate in which $K$ equals the number of ingredients in the target mixture~\cite{feng2026smellnet}. Such a metric is informative about ranking. However, it cannot reveal a wrong ingredient: when a model is told to report exactly $K$ ingredients, every false positive is counted only as a missing hit. It also leaves open why unseen mixtures fail. Plausible explanations include a weak architecture, too few recordings of some ingredients, a bias towards frequent labels and sensor drift between recording sessions~\cite{zhang2026ldac}, and each would call for a different remedy.

We test these explanations directly. From the SmellNet training data we build three folds that hold out ingredient combinations while keeping every ingredient in training. On these folds we train four architectures (MLP, CNN, LSTM and the Transformer encoder of ScentFormer) with three seeds each. We score their presence predictions without telling them how many ingredients a sample contains, report each rank of the prediction separately, check whether predicted pairs reproduce training combinations, and compare failures across architectures and ingredients. After fixing all analyses, we retrain each model on the full training data and evaluate it once on the official unseen split.

The audit separates two abilities that the original metric merges. Every model finds one ingredient of a binary mixture reliably, with first-ranked accuracy of 60--67\%, while second-ranked accuracy is 16--25\%, against 17\% for a random guess. Instead, the second choice tends to be a familiar partner of the first, so that predicted pairs reproduce training combinations about twice as often as chance. Under the count-informed protocol, 90--96\% of validation mixtures contain at least one ingredient that is absent. A single evaluation on the official unseen split confirms the hidden errors and shows a harder picture: there, even the first-ranked ingredient is correct in only 29--46\% of binary mixtures, and on the combinations of that split that are truly absent from training no model names both ingredients in more than 1.2\% of them. The failures are shared. Per-ingredient recall correlates at 0.81--0.97 across the four architectures, and on our folds pear, strawberry and garlic are never placed among the top two by any model, although garlic appears in 110 of the 731 training recordings. Finally, a frequency correction based on logit adjustment~\cite{menon2021logit} raises second-ranked accuracy for every model, but it costs more correct first-ranked ingredients than it gains.

These results locate the bottleneck in the ingredients and in the training distribution rather than in the network. Changing the architecture leaves the same ingredients unrecognised, and the measured signal of those ingredients is barely separable by any model we tested. The models also favour ingredients that are frequent in training, and a frequency correction moves this preference between ingredients instead of removing it. Pure recordings, in turn, split these ingredients further: some are separable alone and lost only in mixtures, while others are hard to separate even alone. Our contributions are:
\begin{itemize}
\item an evaluation of SmellNet mixture recognition without oracle cardinality, which shows that the count-informed metric hides wrong ingredients in 90--96\% of mixtures for four architectures, on internal folds and on the official unseen split (Sections~\ref{sec:oracle} and~\ref{sec:official});
\item a rank-wise analysis showing that, on internal folds, models recognise one ingredient and recall a familiar partner for the second, while on the official split even the first rank is unreliable (Sections~\ref{sec:rank} and~\ref{sec:official});
\item evidence that the failures are shared across architectures, are not explained by recording counts, and are not removed by frequency correction, together with a pure-versus-mixture comparison that separates ingredients lost by mixing from ingredients the sensors barely resolve (Sections~\ref{sec:shared}--\ref{sec:adjust}).
\end{itemize}

\section{Related Work}

\textbf{Mixtures in machine olfaction.} SmellNet~\cite{feng2026smellnet} introduced a large dataset of natural substances and their mixtures recorded with low-cost gas sensors, together with baselines for ingredient presence and mixing ratios. Later work on this data has focused on new models: UnMixNet couples a graph network with a Maxwell--Stefan transport model to estimate mixture composition~\cite{shi2026unmixnet}, and LDAC-Net learns multi-lag temporal differencing to resist sensor drift~\cite{zhang2026ldac}. Earlier electronic-nose work studied mixtures that contain gases unseen in training and found that a target gas is identified reliably only when it dominates the mixture~\cite{fan2022unknown}. In contrast, every ingredient in our setting is known and only the combination is new, and rather than proposing a network we change the evaluation.

\textbf{Compositional generalization.} Benchmarks for compositional generalization hold out combinations while keeping their parts familiar, and they measure how far the distribution of combinations shifts between training and test~\cite{keysers2020cfq}. Hupkes et al.\ call the ability to recombine known parts in new ways systematicity~\cite{hupkes2020compositionality}. In vision, the appearance of an attribute depends on the object it modifies, so that the red of a wine differs from the red of a tomato~\cite{misra2017redwine}. A mixture poses a similar problem in sensor space, since the response to one ingredient is shaped by its partner.

\textbf{Label imbalance in multi-label learning.} Logit adjustment removes the bias of a classifier towards frequent classes by offsetting each logit with its log prior, either after training or inside the loss~\cite{menon2021logit}. Multi-label variants additionally account for label co-occurrence and for the dominance of negative labels~\cite{wu2020dbloss,benbaruch2021asl}. We test the simplest member of this family, a per-ingredient frequency offset, because it is the direct remedy for the frequency explanation; co-occurrence-aware losses target the pairing behaviour that we observe and are a natural next step.

\textbf{Evaluation incentives.} Kalai et al.\ argue that language models produce confident errors partly because common benchmarks reward a guess more than an admission of uncertainty~\cite{kalai2025hallucinate}. A hit rate with a known number of answers creates the same incentive for ingredient recognition, since it never penalises a wrong ingredient that fills an available slot.

\section{Experimental Setup}\label{sec:setup}

\subsection{Data and combination-held-out folds}
We use the mixture part of the SmellNet training data: 731 recordings of 12 ingredients, of which 147 are pure ingredients, 557 binary mixtures and 27 ternary mixtures. Each recording contains four gas channels sampled every 100\,ms. The local release has more recordings than the 679 training sessions reported in the paper~\cite{feng2026smellnet}, and we report results on the release we could inspect.

Sections~\ref{sec:oracle}--\ref{sec:adjust} use no official test data. Instead, we build three internal folds from the training data. Each fold holds out a set of ingredient combinations, called supports, together with every recording of those supports at all mixing ratios, and keeps all pure recordings in training. Every ingredient therefore appears in training, while no validation support does (Table~\ref{tab:folds}). This follows the split principle of compositional benchmarks in which the atoms of the test set are familiar and their compounds are new~\cite{keysers2020cfq}.

\begin{table}[t]
\centering
\caption{Combination-held-out folds. No validation support appears in training.}
\label{tab:folds}
\begin{tabular}{lrrrrr}
\toprule
Fold & Train & Validation & Binary & Ternary & Supports \\
\midrule
1 & 539 & 192 & 178 & 14 & 7 \\
2 & 535 & 196 & 196 & 0 & 7 \\
3 & 535 & 196 & 183 & 13 & 8 \\
\bottomrule
\end{tabular}
\end{table}

\subsection{Models and training}
We train the four architectures released with SmellNet: an MLP, a 1D CNN, an LSTM~\cite{hochreiter1997lstm} and the Transformer~\cite{vaswani2017attention} encoder of ScentFormer, each with the presence head of the released code. We use the released training procedure without changes. Recordings are cut into windows of 50 samples with a stride of 25, so that the first 600 samples (60\,s), which the released loader keeps, yield 23 windows. As in the released code, the inputs contain the four gas channels and the timestamp column, standardised with statistics of the training windows. The timestamp runs from 0 to 59.9\,s within every recording, so it carries no information about the recipe or the recording date. The loss combines a KL term on mixing ratios, an $\epsilon$-insensitive ratio term and a focal binary cross-entropy~\cite{lin2017focal} on ingredient presence, with on-the-fly synthetic mixing of training windows. This augmentation averages two or three training windows with random weights and labels the result accordingly, so it can create synthetic versions of held-out supports, for example from a pure pear and a pure garlic window. Such windows are linear blends of sensor signals rather than recordings of physical mixtures, so the held-out supports remain unseen as measured mixtures; we keep the augmentation because it is part of the released procedure. All models are trained for 60 epochs with Adam (learning rate $10^{-3}$, weight decay $3\times10^{-4}$, batch size 64) and seeds 7, 42 and 123. A recording receives the mean of its window probabilities. Every number in Sections~\ref{sec:oracle}--\ref{sec:adjust} is the mean of the nine fold--seed runs unless stated otherwise. For Section~\ref{sec:official}, we retrain each architecture with the same three seeds on all 731 training recordings and evaluate once on the 213 recordings of the official unseen split, after all other analyses were fixed.

\subsection{Evaluation without oracle cardinality}\label{sec:metrics}
For a binary validation mixture $n$ with ingredient set $S_n$, let $j_n(1)$ and $j_n(2)$ be the two ingredients with the highest presence scores. We report the accuracy of each rank,
\begin{equation}
R_r=\frac{1}{N}\sum_{n=1}^{N}\mathbf{1}\{j_n(r)\in S_n\},\qquad r\in\{1,2\},
\end{equation}
the exact-pair accuracy $E_2$, which counts mixtures with $\{j_n(1),j_n(2)\}=S_n$, and the number of wrong ingredients among the top two, $W_2=2-R_1-R_2$. A random ranking of 12 ingredients gives $R_1=R_2=1/6$ and $E_2=1/66$.

To see whether predictions follow memorised combinations, we compute how often the top-two pair of a binary mixture is a combination that occurs in the training fold. As a reference, we use the fraction of the remaining 11 ingredients that co-occur with the first-ranked ingredient in training, which is the rate expected if the second choice were random. We call the ratio of the two the familiar-pair ratio. Because no validation pair occurs in training, a model that always names the true pair has a ratio of 0, a random second choice gives 1, and values above 1 indicate a pull towards training combinations.

For comparison with the original protocol, we also apply the count-informed rule of SmellNet, which reports the $K$ ingredients with the largest predicted mixing ratios, where $K$ is the true number of ingredients~\cite{feng2026smellnet}. This rule is evaluated on all validation mixtures.

\section{Results}

\subsection{Count-informed evaluation hides wrong ingredients}\label{sec:oracle}
Under the count-informed rule of SmellNet, the four models recover 41--44\% of the ingredients in unseen combinations. The same predictions contain at least one absent ingredient in 89.8\% (CNN) to 96.1\% (LSTM) of validation mixtures, and in no single run of the 36 below 73\%. These two numbers describe one failure. When exactly $K$ ingredients must be reported, each absent ingredient displaces a present one, so a false positive appears in the hit rate only as a missing hit and the count of wrong ingredients is never reported. A model that ranks one correct ingredient first and fills the remaining slots with plausible partners therefore receives partial credit for a set that is wrong in almost every mixture.

\begin{figure}[t]
\centering
\includegraphics[width=\columnwidth]{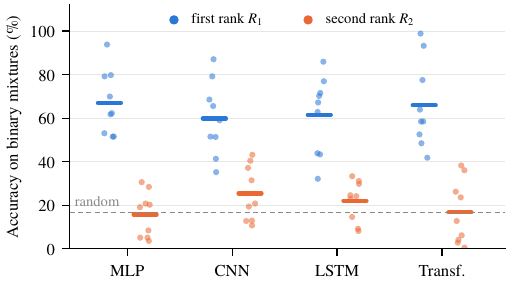}
\caption{First-ranked ingredients are usually present; second-ranked ingredients are present about as often as a random guess. Dots are the nine fold--seed runs, bars their mean, on binary validation mixtures of unseen combinations.}
\label{fig:rank}
\end{figure}

\subsection{One ingredient is found, the second is recalled}\label{sec:rank}
Without the count, the predictions split into two parts of different quality (Fig.~\ref{fig:rank}). The first-ranked ingredient is present in 59.9--67.0\% of binary mixtures, about four times the random rate of 16.7\%. The second-ranked ingredient is present in only 15.7--25.4\%, so the MLP and the Transformer place the second ingredient no better than chance and the CNN, the strongest model on this measure, about 1.5 times better. The gap between the ranks varies across folds, with a standard deviation of 16--27 points, but first-rank accuracy exceeds second-rank accuracy in all nine runs of every model. As a result, models name both ingredients of a pair in 1.5--7.2\% of binary mixtures, where a random ranking would reach 1.5\%.

\begin{figure*}[t]
\centering
\includegraphics[width=\textwidth]{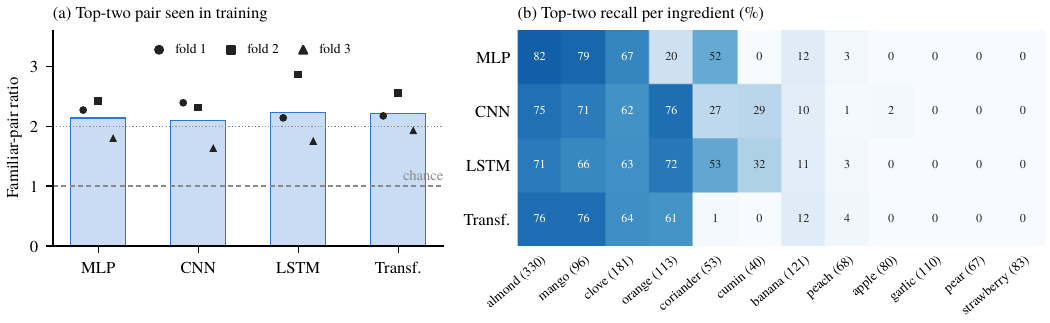}
\caption{Failures follow the training combinations and the ingredients, not the architecture. (a) Ratio between the rate at which the top-two pair of a binary mixture occurs in training and the rate expected from a random second choice; bars pool all runs, markers show folds. (b) Fraction of binary mixtures containing each ingredient in which the ingredient is ranked among the top two; numbers in brackets give the training recordings that contain the ingredient.}
\label{fig:shared}
\end{figure*}

The second choice is not random, however. Predicted pairs are combinations that occur in training 2.09--2.23 times as often as expected from the first-ranked ingredient's training partners (Fig.~\ref{fig:shared}a). The ratio exceeds 2 in folds 1 and 2 for every model and lies between 1.64 and 1.93 in fold 3. Models thus tend to complete a mixture with a partner they have seen next to the first ingredient, rather than with the ingredient that is actually present.

\subsection{Failures are shared across architectures}\label{sec:shared}
The four architectures fail on the same ingredients (Fig.~\ref{fig:shared}b). Their per-ingredient top-two recall correlates at 0.81--0.97 across all six pairs of models. Almond, mango and clove are found in 62--82\% of the binary mixtures that contain them, by every model. Pear, strawberry and garlic are never ranked among the top two, and apple and peach in at most 4\% of their mixtures.

The number of recordings does not explain this split. Garlic occurs in 110 training recordings, more than mango (96), whereas cumin occurs in only 40 and is still found by the CNN and the LSTM in about 30\% of its mixtures. A threshold-free measure points in the same direction. The area under the ROC curve (AUC) for separating validation mixtures that contain an ingredient from those that do not averages 0.52--0.59 over pear, apple, strawberry, garlic and peach, against 0.81--0.87 over almond, mango, clove and orange, depending on the network. For these five ingredients the networks order recordings almost at random, whatever score offset they apply. The signal is not absent for every one of them: a logistic regression on the mean of each gas channel separates apple with an AUC of 0.88 and 0.95 in its two evaluable folds.

Pure recordings, finally, show where the separation is lost. We classify the 147 pure recordings of the mixture data with a logistic regression on channel averages, using session-grouped cross-validation, and compute the same per-ingredient AUC (Table~\ref{tab:pure}). Pear and strawberry are well separated when recorded alone (AUC 0.89 and 0.87) but not in mixtures (0.66 and 0.38), so their failure arises in mixtures. For strawberry the mixture AUC is even below 0.5: the networks score strawberry mixtures lower than mixtures without strawberry, which suggests that its partners or recording sessions mislead the models in addition to masking its response. Peach, garlic and apple remain weakly separated even when pure (0.56--0.70), which points to a limit of the sensor channels or of the features rather than of mixing.

\begin{table}[t]
\centering
\caption{Per-ingredient AUC on pure recordings (logistic regression, ten session-grouped splits) and on mixtures (mean of the four networks).}
\label{tab:pure}
\begin{tabular}{lcc}
\toprule
Ingredient & Pure & Mixture \\
\midrule
pear & 0.89 & 0.66 \\
strawberry & 0.87 & 0.38 \\
apple & 0.70 & 0.65 \\
garlic & 0.68 & 0.55 \\
peach & 0.56 & 0.61 \\
\midrule
almond, mango, clove, orange (mean) & 0.90 & 0.85 \\
\bottomrule
\end{tabular}
\end{table}

\subsection{Frequency correction reallocates errors}\label{sec:adjust}
A bias towards frequent labels is a natural remedy target, and logit adjustment~\cite{menon2021logit} corrects it by subtracting the log prior of each class. We apply its post-hoc form to each ingredient independently, ranking by $\operatorname{logit}(p_i)-\operatorname{logit}(\pi_i)$, where $\pi_i$ is the fraction of training recordings of the fold that contain ingredient $i$, and we fix the temperature $\tau=1$ before seeing any result.

\begin{table}[t]
\centering
\caption{Post-hoc logit adjustment on binary mixtures (mean of nine runs). Second-rank accuracy rises, but more first-ranked ingredients are lost, so the number of wrong ingredients in the top two grows for every model.}
\label{tab:adjust}
\setlength{\tabcolsep}{3pt}\footnotesize
\begin{tabular}{lcccc}
\toprule
Model & $R_1$ (\%) & $R_2$ (\%) & $E_2$ (\%) & $W_2$ \\
\midrule
MLP & 67.0 $\to$ 34.5 & 15.7 $\to$ 29.6 & 1.5 $\to$ 4.0 & 1.17 $\to$ 1.36 \\
CNN & 59.9 $\to$ 42.0 & 25.4 $\to$ 27.7 & 7.2 $\to$ 4.9 & 1.15 $\to$ 1.30 \\
LSTM & 61.6 $\to$ 39.9 & 22.0 $\to$ 26.9 & 4.3 $\to$ 3.8 & 1.16 $\to$ 1.33 \\
Transformer & 65.9 $\to$ 47.6 & 16.7 $\to$ 28.2 & 5.4 $\to$ 3.4 & 1.17 $\to$ 1.24 \\
\bottomrule
\end{tabular}
\end{table}

The correction raises second-rank accuracy for every model, by 2.3 (CNN) to 13.9 (MLP) points, but it lowers first-rank accuracy by more, so the expected number of wrong ingredients among the top two increases for all four (Table~\ref{tab:adjust}). In paired comparisons within a fold and seed, this number rises in 7--9 of the nine runs of each model, whereas second-rank accuracy rises in only 4--8. Exact-pair accuracy improves only for the MLP. The adjustment moves attention from frequent ingredients to rare ones without improving the ranking within either group, which is what the per-ingredient AUC predicts: a constant offset cannot reorder recordings for a single ingredient. Training the MLP with the adjustment inside the presence loss gives the same pattern, with second-rank accuracy of 25.3\% and 1.38 wrong ingredients among the top two.

\subsection{The official unseen split}\label{sec:official}
The official unseen split contains 213 recordings of 16 ingredient combinations, 182 of them binary. In the released data, four of these combinations, all with almond, also occur in training at other mixing ratios; they account for 48 of the binary recordings, so the split is not fully combination-held-out. Its binary mixtures involve only seven ingredients, mostly banana, peach, coriander and garlic, and none of them contains clove. Table~\ref{tab:official} summarises the single evaluation, averaged over three seeds. As a check of our setup, the released code with seed 42 reproduces the dynamic Top-$K$ hit rate reported for the MLP on this split exactly (34.0\%)~\cite{feng2026smellnet}; for the CNN, LSTM and Transformer it gives 27.8--31.2\% against the reported 34.2--38.9\%, a gap that may come from run-to-run variation or from preprocessing options of ScentFormer that we do not use.

\begin{table}[t]
\centering
\caption{Single evaluation on the official unseen split (mean $\pm$ standard deviation over three seeds). Oracle: count-informed rule, share of mixtures with at least one absent ingredient. Ratio: familiar-pair ratio.}
\label{tab:official}
\setlength{\tabcolsep}{2.5pt}\footnotesize
\begin{tabular}{lccccc}
\toprule
Model & Oracle (\%) & $R_1$ (\%) & $R_2$ (\%) & $E_2$ (\%) & Ratio \\
\midrule
MLP & 95.1$\pm$3.5 & 31.9$\pm$5.7 & 28.8$\pm$1.8 & 5.9$\pm$1.8 & 1.56 \\
CNN & 93.7$\pm$4.5 & 33.3$\pm$5.0 & 28.4$\pm$7.3 & 7.5$\pm$4.4 & 1.56 \\
LSTM & 92.5$\pm$1.7 & 28.6$\pm$7.4 & 30.4$\pm$1.8 & 8.6$\pm$2.8 & 1.82 \\
Transformer & 92.0$\pm$1.6 & 45.8$\pm$3.7 & 14.7$\pm$4.7 & 5.1$\pm$3.7 & 1.57 \\
\bottomrule
\end{tabular}
\end{table}

Two findings carry over. The count-informed rule again places an absent ingredient in 92--95\% of mixtures, and the per-ingredient top-two recall again correlates strongly across architectures (0.84--1.00), with coriander and peach almost never found by any model. In contrast, the rank pattern does not carry over. Only the Transformer keeps a clear gap between the ranks; for the other models the first-ranked ingredient is as unreliable as the second, at 29--33\%. A large part of the first-rank errors comes from clove, which no binary test mixture contains but which the models rank first in 22--40\% of them. Clove is the second most frequent ingredient in training, and coriander, which is easy to separate when pure, appears in training only next to almond and orange but in the test split next to four other partners. The models answer this shift with familiar ingredients. Separating the 48 binary recordings of combinations seen in training from the 134 of truly unseen combinations, a descriptive split of the same predictions, sharpens this picture. On the seen combinations the models rank a correct ingredient first in 69--84\% of mixtures and name both in 19--29\%. On the unseen combinations the first rank is correct in only 13--33\%, both ingredients are named in at most 1.2\%, and the familiar-pair ratio is 1.84--2.38, close to our folds (1.56--1.82 over the whole split).

\section{Discussion}

\textbf{What the evaluation should report.} An electronic nose does not know how many ingredients a sample holds, and a wrong ingredient is a visible error. We therefore recommend reporting, next to any ranking metric, the accuracy of each rank, the exact-set accuracy and the number of wrong ingredients per mixture, all without oracle cardinality, together with per-ingredient AUC, which score shifts between ingredients cannot change. Benchmarks with an unseen split should also verify that its combinations are absent from training at every mixing ratio.

\textbf{Where the bottleneck lies.} Four architectures with different inductive biases fail on the same ingredients, and a larger number of recordings does not protect an ingredient from failure. Changing the network is therefore unlikely to close the gap on this data. The pure recordings divide the failing ingredients into two groups. Pear and strawberry are separable alone and lost in mixtures, so a model that represents how a partner masks or distorts their response could recover them. Peach, garlic and apple are hard to separate even alone, so their recognition likely needs more selective sensor channels or better features; the apple result in Section~\ref{sec:shared} shows that features alone can recover part of it. The six-channel pure recordings released with SmellNet cannot settle the channel question, because all of these ingredients are already separable there with four channels. A second limit lies in the training distribution. The models favour ingredients that are frequent in training, most visibly on the official split, where absent clove is often ranked first. A constant frequency offset shifts this preference between ingredients rather than removing it (Section~\ref{sec:adjust}), so training data with more varied partners, or objectives that model label co-occurrence~\cite{wu2020dbloss}, are more promising remedies.

\textbf{Recording time.} The SmellNet session identifiers can be read as base-36 millisecond timestamps. Under this reading, the training recordings span 44.7 days and each recipe was recorded in consecutive batches. As is known for metal-oxide arrays~\cite{vergara2012drift}, the readings drift: the distance between two sessions of one recipe correlates with their time gap (Spearman 0.681) and, for the quarter of pairs recorded furthest apart, approaches the distance between different recipes. Splits should therefore keep neighbouring sessions together, as recommended for temporally structured data~\cite{roberts2017cv}. The drift does not explain which ingredients fail, however: recall falls with time distance when all ingredients are pooled, but this association reverses once each ingredient is analysed separately.

\textbf{Limitations.} Our three development folds hold out seven or eight supports each and informed several of our earlier analyses. The official unseen split was scored with the count-informed metric during an early reproduction of the released code, before any of the analyses reported here; the evaluation in Section~\ref{sec:official} was run once, after the analyses were fixed. The split contains 213 files but only 197 distinct signals, and we score it by file. The pure recordings are grouped by session but not by recording time, which may overstate their separability. The MLP was trained on a CPU and the other models on a GPU. The released data contain more recordings than the paper reports, and the session identifiers are interpreted, not documented.

\section{Conclusion}
Evaluated without knowing how many ingredients a mixture contains, four SmellNet architectures recognise one ingredient of an unseen combination on held-out folds and complete the pair with a familiar partner; on the official unseen split even the first ingredient is often a familiar but absent one. The count-informed hit rate hides wrong ingredients in 90--96\% of mixtures, and the failures concentrate on the same ingredients for every architecture. We expect progress on this task to come from making those ingredients separable, through better sensing or representations, and from training data that pair each ingredient with more partners, rather than from a different network or a frequency correction. Such progress should be measured with set-level errors. Code, fold definitions and model predictions are available from the authors on request.

\IfFileExists{IEEEtran.bst}{\bibliographystyle{IEEEtran}}{\bibliographystyle{ieeetr}}
\bibliography{refs}
\end{document}